\documentclass{article}

\usepackage[final]{nips_2018}
\usepackage{amssymb,amsmath,amsthm,amsfonts}
\usepackage{algorithm}
\usepackage[noend]{algpseudocode}

\theoremstyle{definition}
\newtheorem{definition}{Definition}
\theoremstyle{remark}

\usepackage[suppress]{color-edits}  % use this to suppress the package

\addauthor{sw}{blue}     % sw for Steven
\usepackage[utf8]{inputenc} % allow utf-8 input
\usepackage[T1]{fontenc}    % use 8-bit T1 fonts
\usepackage{hyperref}       % hyperlinks
\usepackage{url}            % simple URL typesetting
\usepackage{booktabs}       % professional-quality tables
\usepackage{amsfonts}       % blackboard math symbols
\usepackage{nicefrac}       % compact symbols for 1/2, etc.
\usepackage{microtype}      % microtypography

\title{Privacy-Preserving Distributed Deep Learning for Clinical Data}

\author{
Brett K. Beaulieu-Jones \\
  Department of Biomedical Informatics\\
  Harvard Medical School\\
  Boston, MA 02130 \\
  \texttt{Brett\_Beaulieu-Jones@hms.harvard.edu}
   \And
  William Yuan \\
  Department of Biomedical Informatics\\
  Harvard Medical School\\
  Boston, MA 02130 \\
  \texttt{william\_yuan@g.harvard.edu}
 \And
  Samuel G. Finlayson \\
  Department of Systems Biology\\
  Harvard Medical School\\
  Boston, MA 02130 \\
  \texttt{samuel\_finlayson@hms.harvard.edu}
   \And
Zhiwei Steven Wu\\
Computer Science \& Engineering\\
University of Minnesota\\
Minneapolis, MN, 55455\\
\texttt{zsw@umn.edu}
}

\begin{document}

\maketitle

\begin{abstract}
Deep learning with medical data often requires larger samples sizes than are available at single providers. While data sharing among institutions is desirable to train more accurate and sophisticated models, it can lead to severe privacy concerns due the sensitive nature of the data. This problem has motivated a number of studies on distributed training of neural networks that do not require direct sharing of the training data. However, simple distributed training does not offer provable privacy guarantees to satisfy technical safe standards and may reveal information about the underlying patients. We present a method to train neural networks for clinical data in a distributed fashion under differential privacy. We demonstrate these methods on two datasets that include information from multiple independent sites, the eICU collaborative Research Database and The Cancer Genome Atlas.
\end{abstract}

\section{Introduction}

Machine learning for healthcare requires balancing the need for large participant populations with the responsibility to maintain the privacy of individual participants. This tradeoff has become particularly apparent in the era of large-scale deep neural networks, which have been shown to be very effective for many healthcare applications but frequently require more data than available at any individual institution \citep{ching2018opportunities}. Efforts such as the eICU Collaborative Research Database and The Cancer Genome Atlas have sought to overcome institutional data limitations by creating centralized repositories of patient data for research. However, centralized solutions generally require time- and resource-intensive deidentification processes that in many cases do not guarantee protection against re-identification. In particular, the addition of rich sources of patient data such as genomic measurements or retinal images make safe deidentification extremely difficult without significantly compromising data utility \citep{beaulieu2017privacy}.

Distributed training provides one strategy for building models across institutions without requiring individual institutions to share their raw patient data externally. However, distributed training itself does not guarantee patient privacy, as common strategies typically involve sharing model weights that can potentially expose information about study participants. \swedit{Motivated by the need to provide formal privacy guarantees, we develop a new distributed training method under \emph{differential privacy}~\citep{DMNS06}, a stringent privacy notion that guarantees that no individual patient’s data has a significant influence on the information released about the
dataset.} In particular, we incorporate differential privacy with \emph{cyclical weight transfer}\swedit{~\citep{jamia}}, by training deep neural networks under differential privacy in a distributed fashion. As a result, bounds on the amount of identifying information belonging to a certain individual can be quantified and minimized.

\subsection{Related work}

\cite{jamia} illustrated the ability to use cyclical weight transfer to train deep neural networks for image classification on two imaging datasets (retinal fundus photographs and mammography images) across four simulated institutions. While a promising proof of concept, this work simulates the different institutions by randomly sampling with equal class distributions. They attempted to add noise via resolution changes and class imbalances, but this simulation does not consider real world institutional biases that may be present in the data. By the same token, the work includes images only, and does not consider other types of clinical data. Most relevantly to this work, \citet{jamia} applied  no privacy measures to the weights. Given that the weights are transferred many times between institutions, this causes potential \swedit{privacy leakage} of the underlying data during training. Furthermore, the lack of privacy guarantees means that even the fully-trained model itself may be queried for membership inference, revealing inclusion of individuals in the training set. \cite{narayanan2008robust}, \cite{el2011systematic}, and \cite{homer2008resolving} illustrate examples of general reidentification attacks and \cite{fredrikson2015model} and \cite{shokri2017membership} demonstrate membership inference attacks first given a machine learning model and then as black box attacks. These attacks demonstrate the necessity to offer privacy protection at the model level.

\swedit{\cite{DP-Deep} developed a method to train deep neural networks under differential privacy, but their initial method required centralized data. \cite{SS15} provided a private collaborative learning protocol. While their protocol also works in a distributed fashion, it requires a central server to coordinate the learning among the participants and heavy communication during training---at each iteration of their gradient descent method, each participant needs to send information of their local gradients to the server. In comparison, our protocol is fully distributed (without any central coordinator) and only requires infrequent communication among the participants. In addition, our work crucially uses the \emph{privacy accountant} developed by \citet{Mironov17}, which allows us to keep track of the cumulative privacy loss through R\'enyi differential privacy---an alternative privacy measure that can also be converted back to standard differential privacy guarantee. This in particular enables us to perform a tighter privacy loss analysis compared to standard composition theorems in differential privacy. }

\section{Data and Methods}
\subsection{Data}
\subsubsection{eICU Collaborative Research Database}
The eICU Collaborative Research Database (\cite{pollard2018eicu}) is a retrospective, deidentified, critical care database organized by Philips Health Care and the MIT Laboratory for Computational Physiology. It is comprised of 208 institutions and currently includes records of 200,859 patient stays. For this study we included patient demographic information (age, gender, ethnicity, weight), information about the admitting unit (type, source), as well as the 50 most common diagnoses and laboratory tests (counts and values) occurring in the first 24 hours after admission. We predict mortality for patients who were discharged or expired after more than 24 hours at the largest five participating sites.

\subsubsection{The Cancer Genome Atlas}
The Cancer Genome Atlas (TCGA) (\cite{brca1}) project aims to construct a comprehensive, publicly-available set of cancer genome, transcriptome, and clinical profiles. Patient data is classified by both cancer type and tissue collection site. For this study, we focused on 994 breast invasive carcinoma (BRCA) patients divided among 38 distinct sites. We created a population-level panel of the 500 genes with the most variant expression, and differentiated the two major subtypes of breast cancer: lobular and ductal. These subtypes differ in terms of treatment response and clinical presentation are normally differentiated based on manual histology review. While differentially expressed genes between the two classes have been identified, the subtypes are not characterized by single sources of variation. Attempts at unsupervised clustering based on gene expression failed to differentiate lobular from ductal cancers (\cite{brca2}).

\subsection{Methods}
\subsubsection{Distributed deep learning with differential privacy}
Algorithm 1 specifies the distributed training procedure with differential privacy and is adapted from the centralized version specified by \cite{DP-Deep}.
\begin{algorithm}
\caption{Training distributed deep learning networks under differential privacy}\label{euclid}
\begin{algorithmic}[1]
\Procedure{Training}{Sites $S$=($s_1, ... s_n$) of datasets $D$=($D_1, ..., D_n$) and models $M$=($m_{1}, ... m_{n}$), parameters: noise value $\sigma$, epochs $E$, batch size $b$, learning rate $\eta$, gradient clip norm  $C = \sigma / b$, loss function $\mathcal{L}(\theta) = \frac{1}{N} \sum_{i} \mathcal{L} (\theta,  x_i$, target privacy budget $(\epsilon_s, \delta_s)$ for each site $s$}
\State Initialize weights $\theta$
\For{$e \in \{1,\ldots , E\}$}
\For{$s \in S$}
%\If{Weights are set}
%\State Set $m_s$weights = Weights
%\EndIf
\For{$t$ in $\{1, \ldots , |D_s|/b\}$} \Comment{For number of batches}
\State Sample a batch $B_t$ of $b$ examples at random from $D_s$ (with replacement)
\For{each $i$ in this $B_t$}
\State compute $g_t(x_i) = \nabla_\theta \mathcal{L}(\theta, x_i)$ \Comment{Compute Per Example Gradient}
\State $g_t(x_i) = g_t(x_i)/ \max\left(1, \frac{\left \Vert g_t(x_i) \right \Vert_2}{C}\right)$  \Comment{Clip the $l_2$ Norm of Gradient}
\State $g_t = \frac{1}{b} (\sum_i g_t(x_i) + \mathcal{N}(0, \sigma^2 C^2 I))$ \Comment{Add noise}
\State $\theta = \theta - \eta \, g_t$ \Comment{Take gradient descent step}
\State{Compute cumulative privacy loss with privacy accountant}
\If{Privacy budget $(\epsilon_s, \delta_s)$ is exhausted}
\State Remove $s$ from $S$
\EndIf
\EndFor
\EndFor
\State{Transfer weights $\theta$ to the next site}
\EndFor

\EndFor
%\State Compute the ($\epsilon$, $\delta$) privacy cost using the privacy accountant.

\State Output final weights as $w$ \Comment{Store the weights for cyclical transfer}
\EndProcedure
\end{algorithmic}
\end{algorithm}

For each training epoch (e), we iterate over each site (s). Within the site we train for number of examples at the site divided by the batch size but choose each batch by randomly sampling with replacement. Within each batch we compute the gradient on a per example basis and then clip the $l_2$ norm of each gradient to establish the maximum bound of any singular gradient. We then add noise proportionate to $\sigma$. We continue training until the loss converges or the specified privacy budget is expended ($\epsilon < 10$,  $\delta = 10^{-5}$),

\subsubsection{Privacy Definition and Privacy Accountant}

Our algorithm satisfies differential privacy, which informally requires that the trajectory of the released output weights throughout the cyclical training process does not reveal much information about any individual patient at any of the sites. More formally:

\begin{definition}[Differential Privacy~\citep{DMNS06}]
Let $\epsilon, \delta\in (0, 1)$. A (randomized) algorithm $\mathcal{A}$ satisfies $(\epsilon,\delta)$-differential privacy if for any pair datasets $D$ and $D'$ that differ by one data record and event $S$ in the output range of $\mathcal{A}$, the following holds
\[
\Pr[\mathcal{A}(D)\in S] \leq \exp(\epsilon) \Pr[\mathcal{A}(D') \in S] + \delta.
\]
\end{definition}

For every individual site, we keep tack of the privacy loss parameters $(\epsilon, \delta)$ over the course of training. We leveraged privacy accountant by~\citet{Mironov17}, which provides tight bounds on privacy loss parameters for applications including private SGD method. The privacy accountant takes the sub-sampling rate of $b/|D_s|$ and the noise rate $\sigma$ as input.

\section{Empirical Evaluation}
\subsection{eICU Application}
We tested our method on the eICU data in four different forms (Table 1), 1.) Central, all data is aggregated together and trained without any privacy constraints, 2.) Central private, all data is aggregated together and trained under differential privacy, 3.) Distributed, data is kept at each institution and training occurs via cyclical weight transfer, 4.) Distributed, data is kept at each institutions and training occurs via cyclical weight transfer under differential privacy. We included the five largest institutions as the training data (N = 27,395) and the next five largest institutions at the testing data (N = 20,117). Privacy was calculated on a per institution-basis in order to define the risk of an institution participating in model training. We required a $\delta$ maximum value of $10^{-5}$ and found an $\epsilon$ value of 2.88  in the central private example ($\sigma = 0.5$, batch size = 100, N = 27,395, epochs = 5) and a maximum of 3.84 in the distributed example (N = 4,328).

\begin{table}[]
\centerline{Table 1. AUROC for eICU Mortality prediction task based on number of institutions.}
\centering
\begin{tabular}{lllll}
Institutions & Central & Central Private & Distributed & Distributed Private \\
1 & 0.743 & 0.683 & 0.738 & 0.662 \\
2 & 0.760 & 0.720 & 0.762 & 0.706 \\
3 & 0.794 & 0.780 & 0.789 & 0.753 \\
4 & 0.804 & 0.785 & 0.797 & 0.780 \\
5 & 0.808 & 0.789 & 0.801 & 0.792
\end{tabular}
\end{table}

\subsection{TCGA Application}
TCGA data was trained using similar forms as the eICU data. RNAseq expression profiles for 994 BRCA patients were min-max normalized and the 500 genes with the highest variance in expression were used to create a panel. Only expression profiles corresponding to tumor samples were included, and profiles from patients with more than one tumor sample were averaged together. Composite sites were created based on TCGA-provided Tissue Site Codes. For this analysis, patients from Tissue Sites with codes starting with 'A' or 'B' were assigned to Sites 2 and 3 respectively, while remaining patients were assigned to Site 1.

This analyses included low sample counts and provided a challenging domain for method evaluation (Table 2). We required a $\delta$ maximum value of $10^{-5}$ and found an $\epsilon$ value of 5.87 in the single site central private example, 5.25 for all sites and a maximum $\epsilon$ value of 6.11 in the distributed example. The largest $\epsilon$ value for the distributed example is found at the smallest participating site, site \#3.

\begin{table}[]
\centerline{Table 2. AUROC for TCGA classification task based on number of participating sites.}
\centering
\begin{tabular}{lllll}
Sites & Central & Central Private & Distributed & Distributed Private \\
1 & 0.657 & 0.644 & 0.650 & 0.643 \\
2 & 0.715 & 0.723 & 0.719 & 0.701 \\
3 & 0.767 & 0.751 & 0.761 & 0.744 \\
\end{tabular}
\end{table}

\section{Discussion and Future Work}

In this paper, we provide a differentially private distributed learning protocol that leverages the idea of cyclical weight transfer. Compared to prior work \cite{SS15}, our method
removes the need of a centralized coordinator and significantly reduces the frequent communication among different sites.
Our empirical evaluation demonstrates it is feasible to deploy a distributed machine learning protocol that enjoys a provable differential privacy guarantee, without making large performance sacrifices.

There are several interesting future directions. First, since the data at each site might be drawn from a different distribution, we can potentially target slightly different models to each site. We can consider a hybrid training method that trains a shared representation with data across all the sites, and also trains a "fine-tuning" mapping for each individual site to capture local trends and biases \citep{beaulieu2016semi}. Second, we need to handle the case in which an individual site might have only a small amount of data (e.g. N < 100) where it is difficult to extract useful information from the site under differential privacy. Small samples sizes are increasingly relevant given added dimensionality (pharmacogenetics etc.). Many problems in healthcare suffer from significant class imbalance, but each of the common methods for controlling this have privacy implications. We would like to develop ways to control for this without privacy implications or in a manner where it can be easily accounted for.

\section{Acknowledgements}

BBJ was supported by NLM grant T15LM007092, WY was supported by the NVIDIA Graduate Fellowship Program and SGF was supported by training grants T32GM007753. The content is solely the responsibility of the authors and does not necessarily represent the official views of the National Institute of General Medical Sciences or the National Institutes of Health.

\bibliographystyle{plainnat}
\bibliography{refs.bib}
\end{document}